\documentclass[default,iicol]{sn-jnl}
\jyear{2021}
\theoremstyle{thmstyleone}

\theoremstyle{thmstyletwo}

\theoremstyle{thmstylethree}

\begin{document}

\title{Optimization of Rocker-Bogie Mechanism using Heuristic Approaches}

\author*[1]{\fnm{Harsh} \sur{Senjaliya}}\email{harsh.senjaliya.hs@gmail.com}

\author[2]{\fnm{Pranshav} \sur{Gajjar}}\email{19bce060@nirmauni.ac.in}

\author[3]{\fnm{Brijan} \sur{Vaghasiya}}\email{vaghasiyabrijan@gmail.com}

\author[4]{\fnm{Pooja} \sur{Shah}}\email{pooja.shah@nirmauni.ac.in}

\author[5]{\fnm{Paresh} \sur{Gujarati}}\email{paresh.gujarati@utu.ac.in}

\affil*[1]{\orgdiv{Department of Mechatronics}, \orgname{Uka Tarsadia Universiy}, \orgaddress{\street{Maliba Campus}, \city{Bardoli}, \postcode{394350}, \state{Gujarat}, \country{India}}}

\affil[2]{\orgdiv{Department of Computer Science and Engineering}, \orgname{Nirma University}, \orgaddress{\street{} \city{Ahmedabad}, \postcode{382481}, \state{Gujarat}, \country{India}}}

\affil[3]{\orgdiv{Department of Mechatronics}, \orgname{Uka Tarsadia University}, \orgaddress{\street{Maliba Campus}, \city{Bardoli}, \postcode{394350}, \state{Gujarat}, \country{India}}}

\affil[4]{\orgdiv{Department of Computer Science and Engineering}, \orgname{Nirma University}, \orgaddress{\street{} \city{Ahmedabad}, \postcode{382481}, \state{Gujarat}, \country{India}}}

\affil[5]{\orgdiv{Department of Mechanical}, \orgname{Uka Tarsadia University}, \orgaddress{\street{Maliba Campus}, \city{Bardoli}, \postcode{394350}, \state{Gujarat}, \country{India}}}

\abstract{Optimal locomotion and efficient traversal of extraterrestrial rovers in dynamic terrains and environments is an important problem statement in the field of planetary science and geophysical systems. Designing a superlative and efficient architecture for the suspension mechanism of planetary rovers is a crucial step towards robust rovers. This paper focuses on the Rocker-Bogie mechanism, a standard suspension methodology associated with foreign terrains. After scrutinizing the available previous literature and by leveraging various optimization and global minimization algorithms, this paper offers a novel study on mechanical design optimization of a rover’s suspension mechanism. This paper presents extensive tests on Simulated Annealing, Genetic Algorithms, Swarm Intelligence techniques, Basin Hoping and Differential Evolution, while thoroughly assessing every related hyper parameter, to find utility-driven solutions. We also assess Dual Annealing and subsidiary algorithms for the aforementioned task while maintaining an unbiased testing standpoint for ethical research. Computational efficiency and overall fitness are considered key valedictory parameters for assessing the related algorithms, emphasis is also given to variable input seeds to find the most suitable utility-driven strategy. Simulated Annealing (SA) was obtained empirically to be the top performing heuristic strategy, with a fitness of 760, which was considerably superior to other algorithms and provided consistent performance across various input seeds and individual performance indicators.
}


\maketitle

\section{Introduction}\label{sec1}

The efficient traversal of a given mobile robot in varying environments, the type of locomotion used plays an essential role. This paper aims to build on the existing technologies and provide a novel comparative analysis for the task of design optimization of a planetary rover’s suspension mechanism.
Due to the recent advents in the field of robotics, and the translation of fixed base robots to autonomous systems, interpreting the optimal method of locomotion that is robust to terrain differences is key for the accurate functioning of a planetary rover \cite{bib1}. The task of improving a rover’s functionality can be improved by using optimization paradigms in conjunction with a rigorously developed objective function that takes into consideration the simulatory data for the rover's associated commissions. With the advent of computational technologies, predictive analysis \cite{bib2}, and the multi-objective use of optimization \cite{bib3}, various algorithms can be leveraged for the said problem statement. The design optimization task can be treated as a numeric global minimization or maximization problem due to the multivariate nature of the fitness equation \cite{bib1}. The paper \cite{bib1} leveraged a Genetic Algorithm for the said task and formulated an objective function. Due to the advancements and refinements presented in the recent literature, extensive experiments can be conducted by improving the fitness constraints and the related optimization algorithms, which work for maximizing fitness. The article \cite{bib4} depicted Particle Swarm Optimization for a related task and proved successful results, this paper also motivated and justified the direction of comparing PSO with Genetic algorithms.
As the Mars Rover is a mobile robot, the wheel suspension system of the rover is most crucial. It allows for movement, mobility, and stability of the robot while it is traveling through a Mars environment. The rover must be able to traverse over obstacles of at least half its wheel diameter and keep its stability on slopes or another rough or hazardous terrain. The rover must be dependable since it must endure dust, corrosion, strong winds, and drastic temperature variations. The majority of rovers are powered by batteries that are replenished during the day by solar panels. As a result, the rover must orient itself so that the solar energy obtained is maximized. The rover’s movement mechanism is critical for it to achieve its goals, perform experiments, collect data, and position itself. In general, there are three forms of rover movement: wheeled, legged, and caterpillar locomotion  \cite{bib31} \cite{bib32}. The primary distinction between various types of planetary robots is the sort of locomotion system. Even though many-legged and hybrid robots have been described in the literature, most researchers continue to concentrate on wheeled locomotion for rovers. The rocker-bogie mechanism is the preferred design for the suspension system of a wheeled planetary rover. This process has been given in several forms in the literature. The suspension arrangement in the Rocky7 Rover \cite{bib33} is identical, but only the front wheels are guided. The Nomad \cite{bib34} use a rocker-bogie system with four steerable wheels hung from two bogies. To overcome obstacles, the CRAB Rover \cite{bib35} uses two parallel bogie systems on each side. The article \cite{bib7} optimize a simplified quasi-static model of the Shrimp, a six-wheeled rover. The motor torques are tuned to decrease wheel slip, which reduces odometric error and power consumption. Given an initial solution, the optimization finds an optimum in the confined solution space \cite{bib23}. The paper \cite{bib14} develop a mathematical model to optimize rover suspension settings that dictate the rocker-bogie shape. The goal is to reduce energy consumption, vertical displacement of the rover’s center of mass, and pitch angle. A sequential quadratic programming algorithm is used by the authors. The paper \cite{bib7} propose a method for optimizing individual wheel ground contact forces. The goal is to maximize traction while minimizing power usage.

\section{Fitness Parameters}\label{sec2}
\subsection{Stability}\label{subsec1}

\begin{figure*}[ht]
\centering
\includegraphics[width=0.46\textwidth]{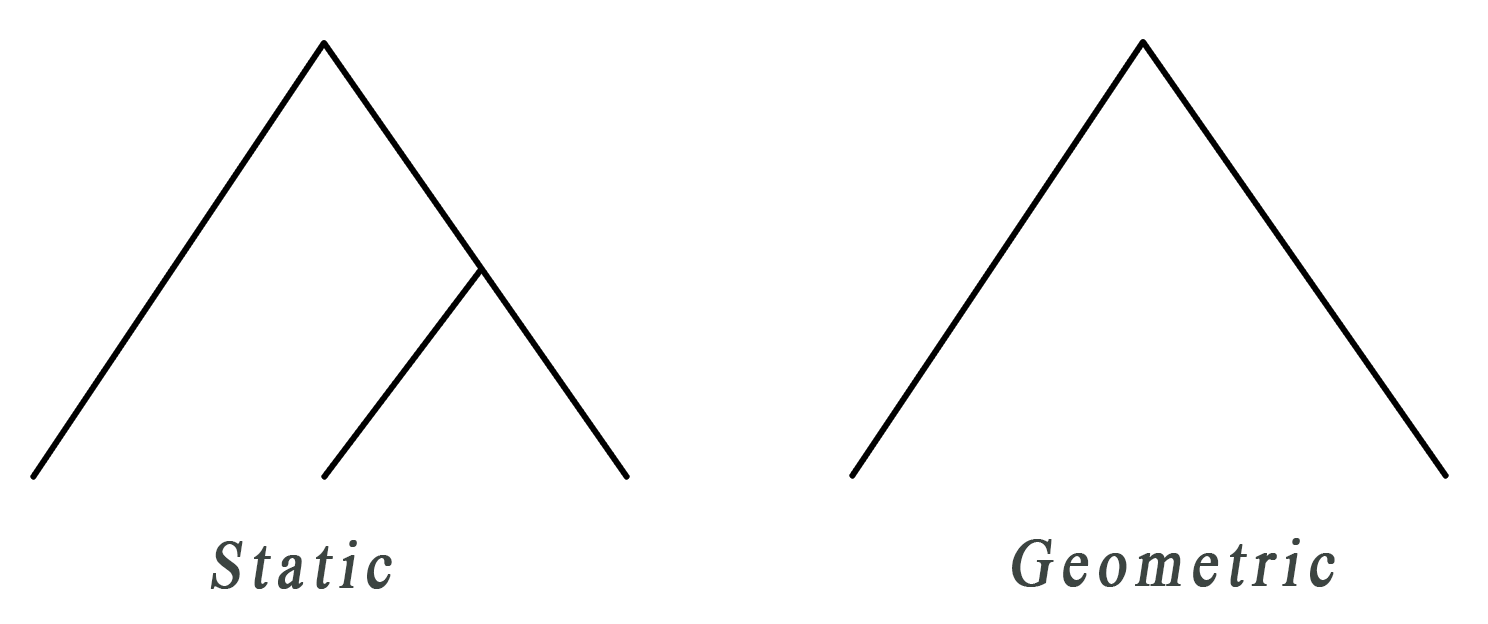}
\caption{Model comparison for evaluating stability \cite{bib5}.}\label{fig1}
\end{figure*}

Stability is one of the essential factors either to plan trajectories, generate velocity commands, or monitor the stability margin online. Depending on the velocity of the rover, there are two methods used to calculate stability can be static or dynamic. In this paper, we have used Static Stability (SS) to calculate the stability of the rover \cite{bib5}. Further, Static Stability (SS) is classified into two approaches Static and Geometric approaches as illustrated in Fig\ref{fig1}.  

\begin{figure*}[ht]
\centering
\includegraphics[width=0.46\textwidth]{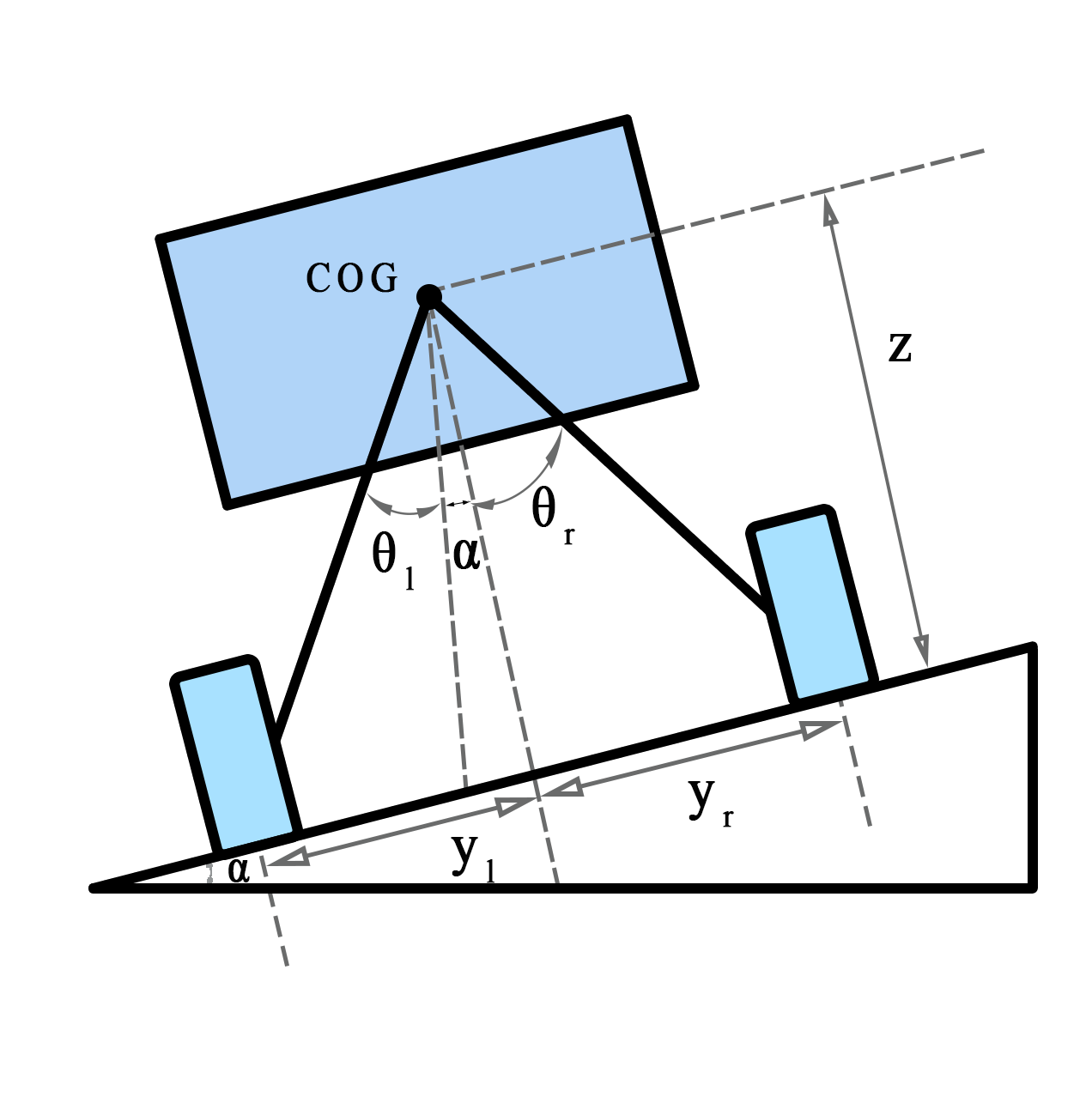}
\caption{Lateral stability of the rover \cite{bib5}.}\label{fig2}
\end{figure*}

\begin{equation}
\theta_{SS}=\alpha tan\left ( \frac{y_{rear}}{z}\right)
\end{equation}

Lateral stability is calculated by calculating the smallest allowable angle on the slope before the rover flips over. If this angle is less than the maximum angle of inclination on the slope at the wheel-terrain contact sites, lateral stability is assured. Geometrically, the angles $\theta_{l}$ and $\theta_{r}$ are obtained as shown in Fig\ref{fig2}.

\begin{figure*}[ht]
\centering
\includegraphics[width=0.49\textwidth]{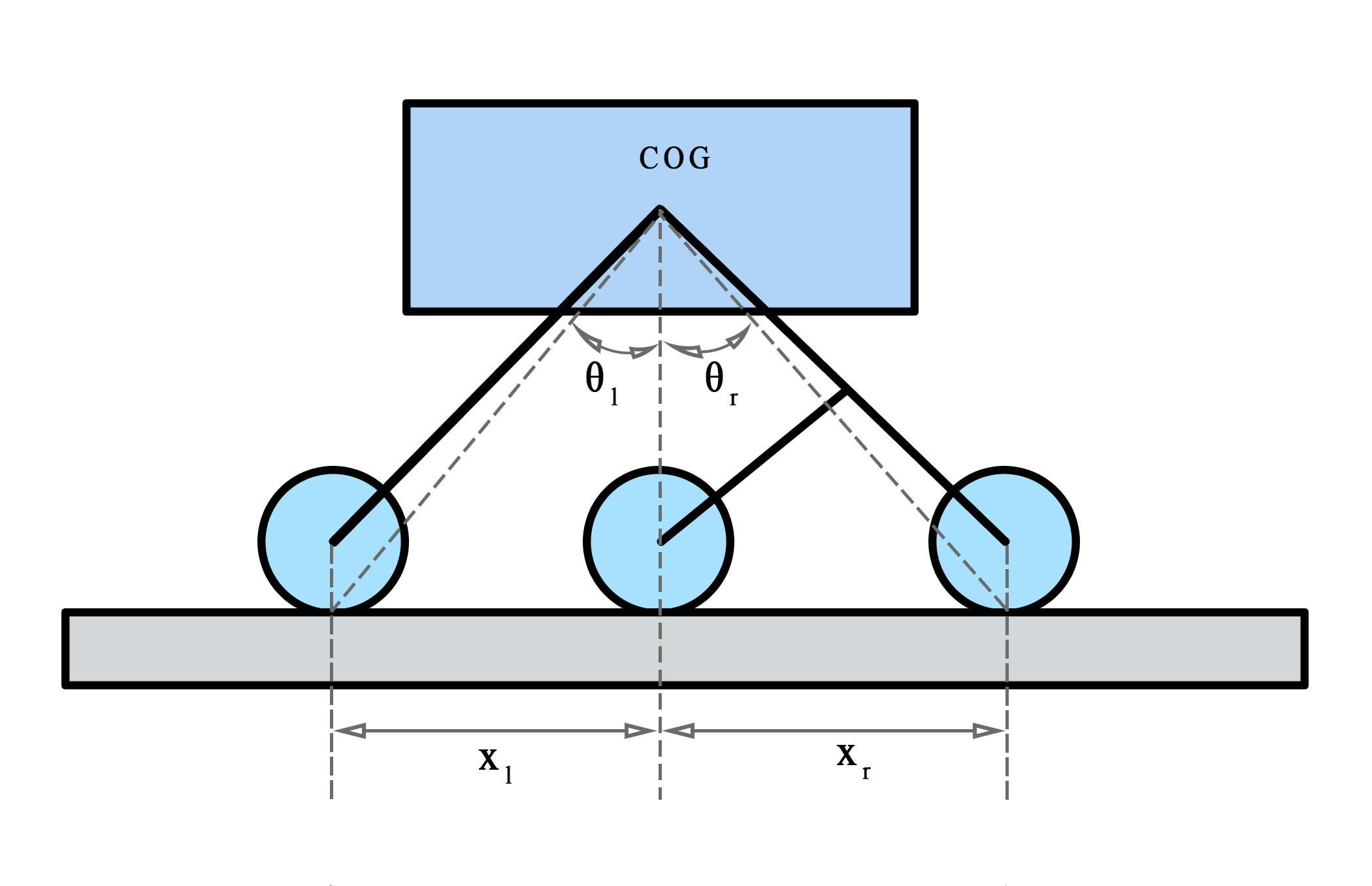}
\caption{Longitudinal stability of the rover \cite{bib5}.}\label{fig3}
\end{figure*}

\begin{equation}
\theta_{SS}=\alpha tan\left ( \frac{x_{rear}}{z}\right)
\end{equation}

The vehicle has longitudinal stability when all wheels make ground contact and the condition $N_i > 0$ is met, where $N_i$ is the normal force at $i^{th}$  wheel which can be seen in Fig\ref{fig3}. The total stability angle $\theta_{stab}$  may be calculated using the formula. \\
 $\theta_{stab} = min ( \theta_r,\theta_l)$\\    
Hence, if the total stability is $\theta_{stab} \geq \alpha$, then the rover's lateral stability is achieved.

\subsection{Power Consumption }\label{subsec2}

Since extraterrestrial rovers are powered by solar energy and should not require recharging or battery replacement throughout the operation, the rover's power consumption must be kept to a minimum. The power consumed by a DC motor-driven wheeled rover employing PWM (Pulse Width Modulation) amplifiers may be determined by the power dissipation in the motor resistances, which can be used to build an optimization criterion for minimal power consumption \cite{bib36}. The rover's power consumption is proportional to the motor torque represented as,

\begin{equation}
{P={\frac {{R}_{m}{{g}^{2}_{m}}} {{K}^{2}_{t}}}\sum^{n}_{i=1} {}}{{ {\tau }}^{2}_{i}}  
\end{equation}

where  $R$ is the resistance of the motor, $K_t$ is the torque constant of the motor, $n$ is the gear ratio of the motor, and $ \tau $ is the torque applied by the $i_{th}$ wheel. Considering \cite{bib6} the traction force $T_i$ may therefore be related to the power usage and $r$ is the wheel radius. 

\begin{equation}
{P={\frac {{R}_{m}{{g}^{2}_{m}{{r}^{2}}}} {{K}^{2}_{t}}}\sum^{n}_{i=1} }{{T}^{2}_{i}} 
\end{equation}

On flat terrain with low wheel sinkage, assuming that the tractive force is equal to the product of the applied wheel torque and the wheel radius. As a result, the control algorithm should strive to decrease $P$ in order to reduce power consumption.

\subsection{Traction and Slip }\label{subsec3}

\begin{figure*}[ht]
\centering
\includegraphics[width=0.46\textwidth]{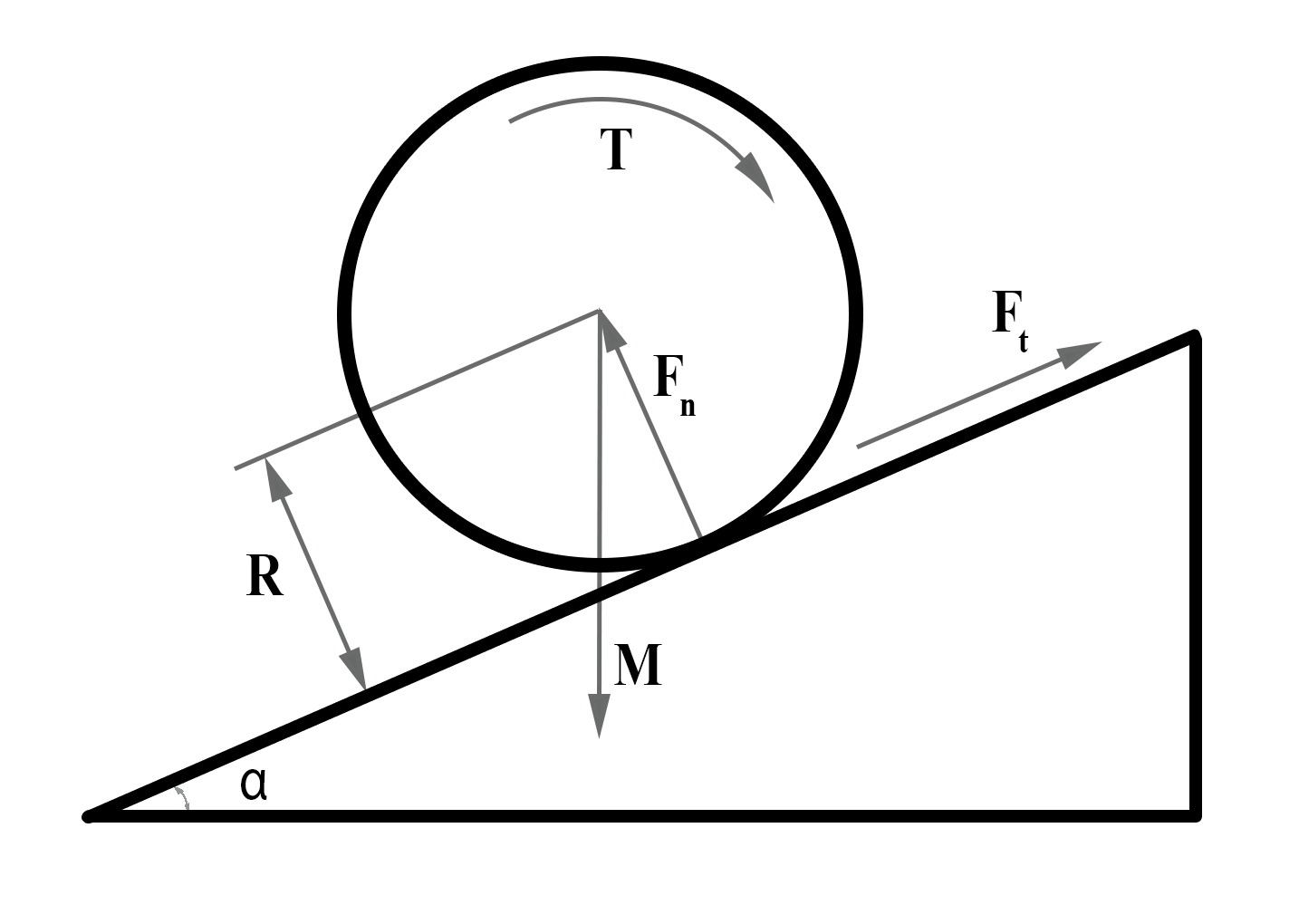}
\caption{Wheel terrain interaction \cite{bib5}.}\label{fig4}
\end{figure*}

The force that generates movement between a body and a tangential surface is known as traction force. The rover must maintain appropriate wheel traction in tough terrain. The rover will be unable to climb over obstacles or steep slopes if traction is too low. When traction is absurdly high, the vehicle must use a significant amount of energy in order to overcome the force and move as shown in Fig\ref{fig4}. Slip occurs when the traction force at a point of contact between a wheel and the ground exceeds the product of the normal force and the friction coefficient.

As a result, if the condition $T_{i} \leq \mu N_i$ is met, no-slip occurs. In actuality, determining the correct friction coefficient for the interaction of two surfaces is quite difficult. As an approximation of the friction coefficient, a virtual friction coefficient is computed as an approximation of the friction coefficient, according to \cite{bib7}, as the ratio of traction to normal force at a single point of contact between a wheel and the ground.

\begin{equation}
{\mu^{ \ast }= max_i \left \{\frac{T_i}{N_i}\right\} }
\end{equation}

The parameter $\mu$ is analogous to a friction coefficient. No-slip happens if ${\mu^{ \ast } \leq \mu} $ the condition is met. As a result, $\mu^{\ast }$ is intended to be minimized. According to \cite{bib7} \cite{bib8}, the optimum solution is found if $\mu^{\ast }$ is equal for all $n$ wheels.

The rear wheels have the greatest chance of slipping while climbing the obstacle. This slippage results in a lack of traction force transfer from the rear to the front wheels. As a result, all of the wheels slips should be minimized. Because the focus of this study is on front leg-wheel design, we assume the remaining five rear wheels are on a level surface, each generating traction force $T_{i}$. The virtual friction coefficient ${\mu^{\ast }= \left \{\frac{T_i}{N_i}\right\} }$ is defined to analyze the possibility of wheel slippage in order to avoid wheel slip.

\subsection{Sinkage}\label{subsec4}

Wheel sinkage is defined as the distance between the lowest point of the wheel in the soil and the horizontal flat ground. Sinkage can occur as a rover travels over uneven terrain, depending on the qualities of the soil as well as the weight, geometry, and form of the wheels. The geometry of the vehicle is optimized in order to decrease the maximum sinkage of the wheels in the terrain. Considering \cite{bib9} an experimentally derived formula for the maximum sinkage experienced by a rigid wheel in weak soil.

\begin{equation}
{ z_{rw} =  \left( \frac{3 W_{w} cos  \theta }{ \big(3-n\big) \big( k_{c}+ b_{w} k_{ \phi }   \big) \sqrt{ d_{w} }   } \right )^{ \frac{2}{ \left(2n+1\right) } } }
\end{equation}

\begin{figure*}[ht]
\centering
\includegraphics[width=0.46\textwidth]{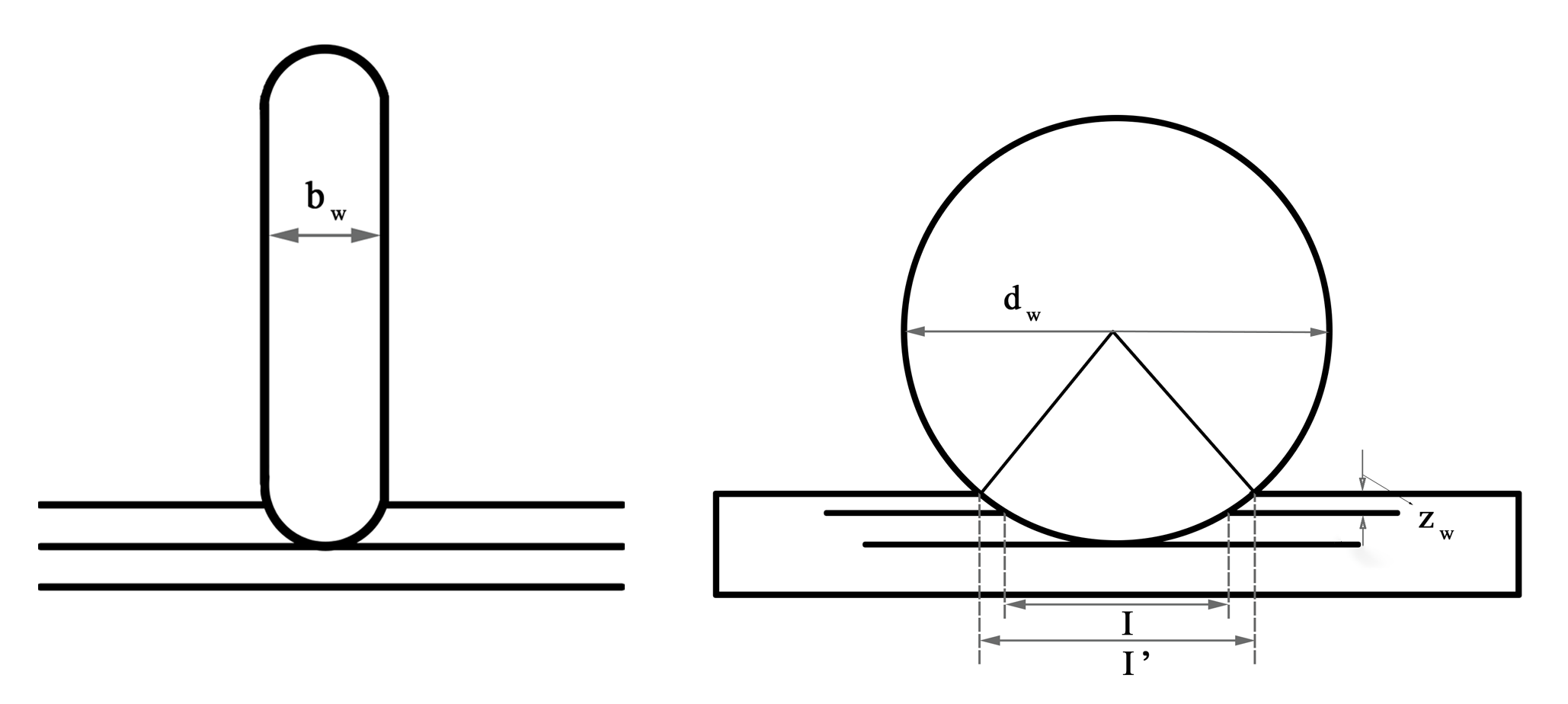}
\caption{Sinkage model of a rigid rolling wheel in soft soil \cite{bib9}.}\label{fig5}
\end{figure*}

Wong \cite{bib10} compiled soil attributes for several soils and used them to describe the soil. To compute the sinkage of the rover's wheel, we selected dry sand as the soil type.

\begin{table*}[ht]
\begin{center}
\begin{minipage}{174pt}
\caption{ Geophysical property of dry sand \cite{bib10}.}\label{tab1}%
\begin{tabular}{@{}lllllll@{}}
\toprule
Soil Type & Moisture Content  & $n$ & $K_c$ &$K_\phi$ & $c$ & $\phi$\\
\midrule
Dry Sand    & 0\%  &  1.10  & 0.1 &3.9 &0.15 &$28^{\circ}$\\

\botrule
\end{tabular}

\end{minipage}
\end{center}
\end{table*}
According to the aforementioned in Fig\ref{fig5} load per wheel is denoted as $W_w$, $K_c$ is cohesive modulus soil deformation,$K_\phi$ is the frictional modulus of soil deformation, $n$ is an exponent of soil deformation, $c$ is coverage of wheel, $d_w$ is diameter and $b_w$ is breath of wheel. 

\subsection{Pitch Variation }\label{subsec5}

The chassis is vital in maintaining the average pitch angles of both rockers by allowing both rockers to move as needed. According to the acute design, one end of a rocker has a driving wheel and the other end is pivoted to a bogie, which provides the requisite motion and degree of freedom \cite{bib11}.
The pitch angle of the chassis $\theta$ where $\gamma$ in the equation is the inclination angle of the rocker linkage. 

\begin{equation}
\displaystyle{ \theta=\frac{\gamma_{1} + \gamma_2}{2}}
\end{equation}

The rover's pitch angle should have very little variation. A differential mechanism connects the two rocker-bogies. As it goes over rough terrain, the rover's pitch is the average of the two rocker deflections \cite{bib12}.

\begin{figure*}[ht]
\centering
\includegraphics[width=0.50\textwidth]{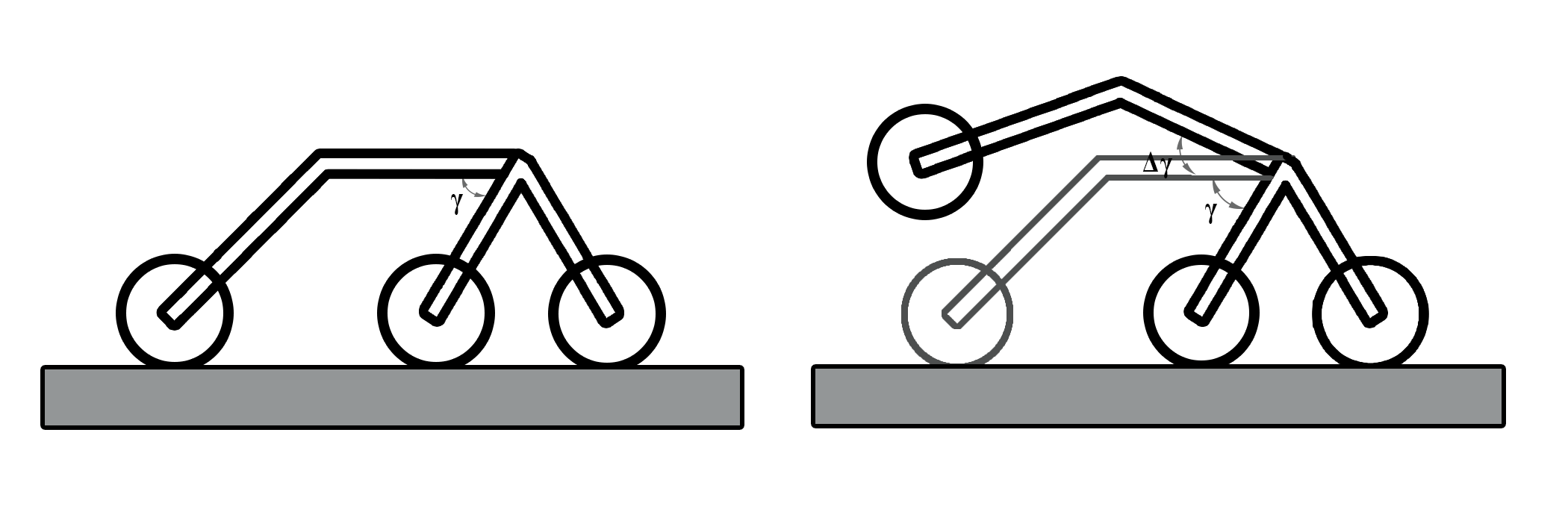}
\caption{Change in pitch angle of  Rocker-Bogie.}\label{fig6}
\end{figure*}

\subsection{Geometric Trafficability }\label{subsec6}
\begin{figure*}[ht]
\centering
\includegraphics[width=0.46\textwidth]{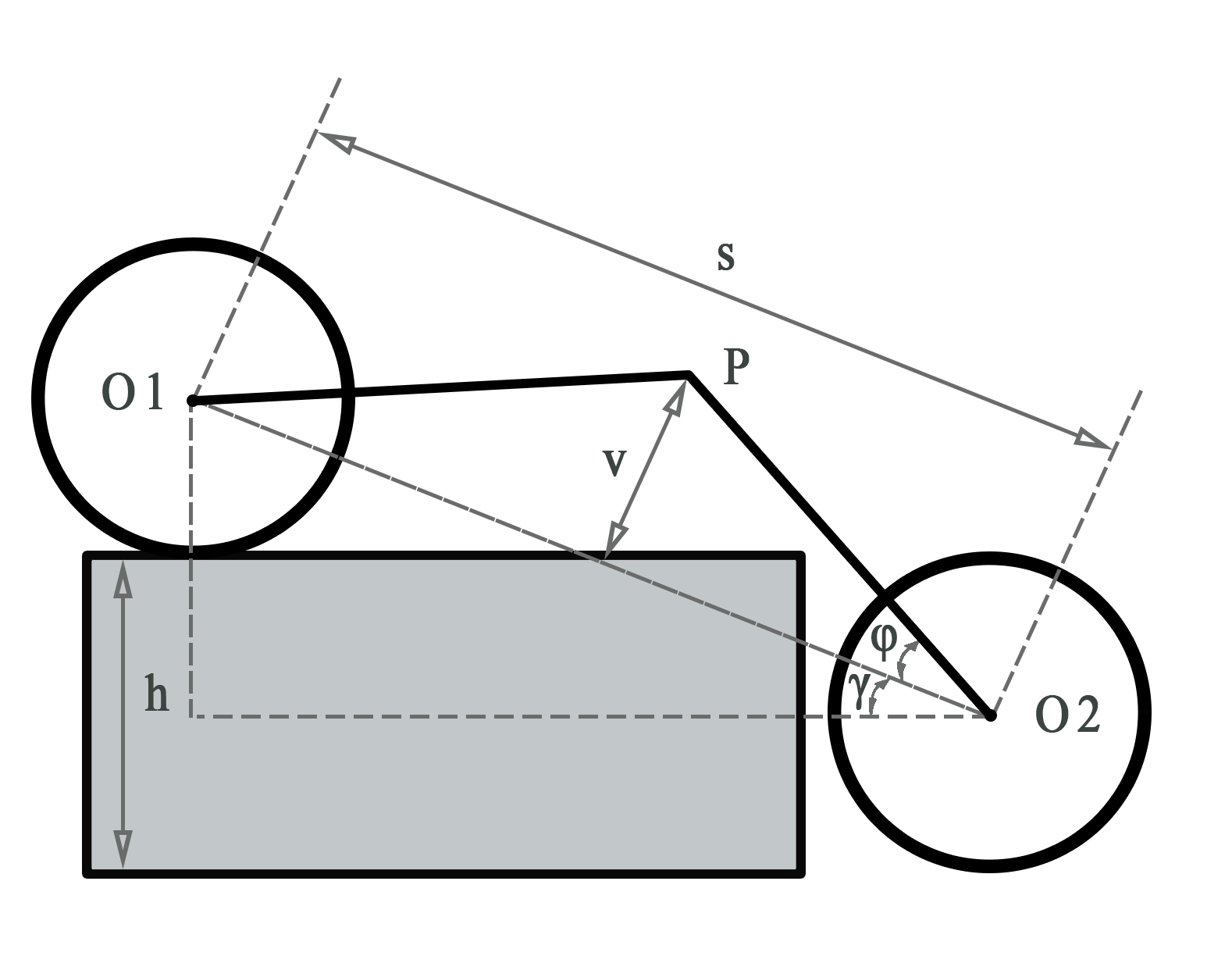}
\caption{The Geometric Trafficability of the Bogie \cite{bib1}.}\label{fig7}
\end{figure*}

The capability of the terrain under consideration to offer mobility for a specific set of vehicles is referred to as trafficability. The ability to anticipate trafficability is dependent on a combination of vehicle's related criteria as well as those linked to terrain cover and substrate material \cite{bib13}. Inadequate geometric trafficability became a key cause of mobility loss. The terrain would not readily impede with the rocker because its swinging scope is significantly larger than that of the bogie \cite{bib14}. The bogie is subjected to a geometric trafficability analysis. The terrain geometry size and suspension parameters, which are primarily the position of the point P and the ground clearance $c$, determine the geometry trafficability of the bogie. When the bogie passes the obstacle with height $h$ and the rear wheel just makes contact with the barrier, the position of the bogie simply maintains in touch with the step for the least constrained position, as illustrated in Fig\ref{fig7}.

\begin{equation}
\displaystyle{z_{t}= \frac{s}{ \sqrt{   \left(\frac{r \sqrt{ s^{2}-h^{2} } + \big(h-r\big)h }{ \big(h-r\big)  \sqrt{ s^{2} - h^{s}} -hr } \right ) ^{2} +1} } }
\end{equation}

According to \cite{bib1}, \cite{bib7}, $z_t$ is the maximum vertical distance between the center of the wheels and the edge of the barrier, where $r$ is the wheel radius and $h$ and $s$ are geometric factors. If the parameter  $z_t$ is greater than or equal to the rover's ground clearance, geometric trafficability is achieved.

\subsection{Load Equalisation}\label{subsec7}

\begin{figure*}[ht]
\centering
\includegraphics[width=0.46\textwidth]{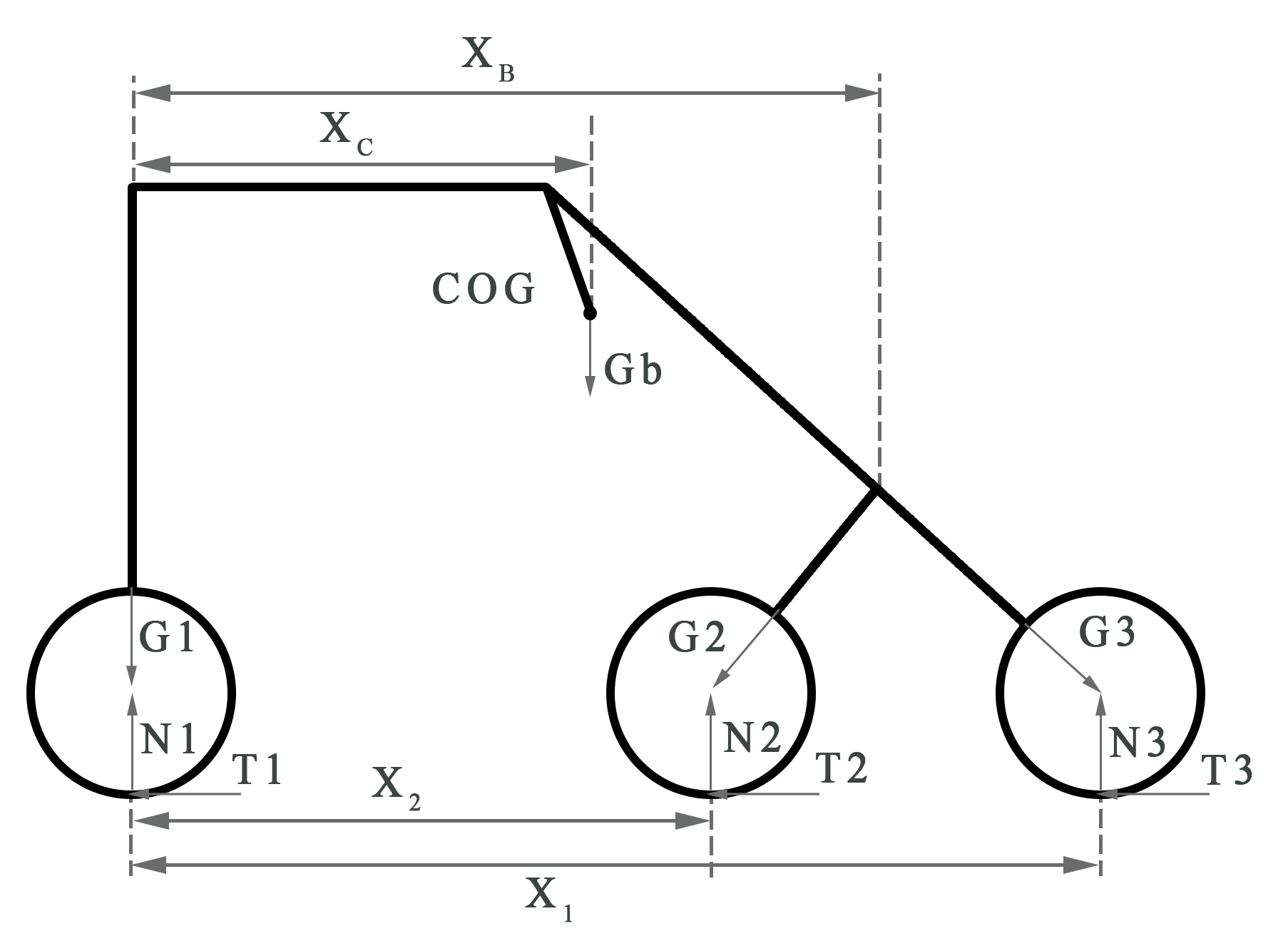}
\caption{Load Equalisation on $n$ Wheels \cite{bib1}.}\label{fig8}
\end{figure*}

Equal load distribution across all $n$ wheels ensures that all wheels have the same working conditions. Fig\ref{fig8} depicts the rocker-bogie suspension's two-dimensional mechanics concept. The entire weight of the rover body is $G_b$, whereas the proportional weight from the rocker-bogie mechanism operating on wheel $i$ is $G_i$. As indicated by \cite{bib14}, a quasi-static force balancing is performed, and the conditions ${x_{b} = 3/2( x_c)}$ and ${ x_c = \frac{1}{2} (x_1 + x_2)}$ are derived from load equalization. Weights $G$, normal forces $N$ and traction forces $T$ are shown in Fig\ref{fig8}. To maximize load equalization, certain geometrical constraints for the suspension system are applied. This study emphasises on the latter condition which correlated to $x_c$.

\section{Optimization Algorithms}\label{sec3}

With the advent of computing technologies, optimization has become an integral part of computer-aided design activities and can be leveraged for the proposed task. An optimization algorithm can be recapitulated as a process that is executed iteratively by comparing various solutions until an optimum or satisfactory solution is encountered \cite{bib15}. For optimizing the overall design of a Rocker Bogie mechanism, a fitness function, and the corresponding additive inverse, the objective function is formulated and minimized using various related algorithms. This paper leverages various heuristic algorithms, which are explained in further subsections.

\subsection{Particle Swarm Optimization}\label{subsec8}

Particle swarm optimization (PSO) \cite{bib16} is a basic bio-inspired technique for searching for an optimal solution in the solution space. It differs from other optimization methods in that it requires only the objective function and is not reliant on the gradient or any differential form of the goal, with fewer hyperparameters \cite{bib16}, \cite{bib37}, \cite{bib38}.  As portrayed in the original paper, sociobiologists consider a flock of birds or a school of fish that moves in a group “can profit from the experience of all other members” \cite{bib17}. In other words, if a bird is flying about aimlessly looking for food, all of the birds in the flock may share their discoveries and assist the entire flock to have the most significant hunt \cite{bib39}.

A flock of birds is a fantastic illustration of animal collective behavior since they fly in large numbers and seldom crash with each other. A flock moves smoothly and is coordinated as though it is commanded by something other than the flock's leader. A flock of birds is a swarm intelligence model, and the birds in it follow particular rules and regulations \cite{bib17}.
The regulations are as follows:
\begin{enumerate}[1.]
\item Every bird makes an effort to avoid colliding with other birds.

\item Every bird goes in the direction of the closest bird.

\item Birds attempt to keep an equal spacing between themselves.

\item A bird communicates with its neighbors. 
\end{enumerate}
Agents are particles in the optimization task parameters space in the PSO. Particles have a location and a velocity vector on each iteration. The relevant objective function value for each position of a particle is determined, and depending on that value, a particle changes its position and velocity according to particular regulations. PSO is a method of stochastic optimization. It does not update current populations but instead operates with a single static population whose members continuously improve as they learn more about the search space. The position updates and the velocity updates are carried out using the following equations (9), (10) respectively. The term $X$ implicated the position and $v$ implicates velocity corresponding to the numeric $t$, $Gbest$ implicates the entities associated with the best performing particle in a generation, and $C$ and $\omega$ are constants.

\begin{equation}
\displaystyle{X_{P} \big(t\big) = X_{t-1}+ v_{P} \big(t\big)}
\end{equation}

\begin{equation}
\begin{split}
X_{P} \big(t\big) = X_{t-1}+ v_{P} \big(t\big)
 v_{p} \big(t\big) =  \omega  \ast v_{P} \big(t\big) \cr + C_{1} \ast rand() \ast \big[ X_{Pbest} -  X \big(t-1\big) \big] \cr +  C_{2}  \ast rand () 
\ast   \big[ X_{Gbest} - X \big(t-1\big) \big]
\end{split}
\end{equation}
     
For a better understanding of PSO, a graphical description of the steps incorporated for reaching optimality is mentioned below in figure \ref{fig9}.

\begin{figure*}[ht]
\centering
\includegraphics[width=0.17\textwidth]{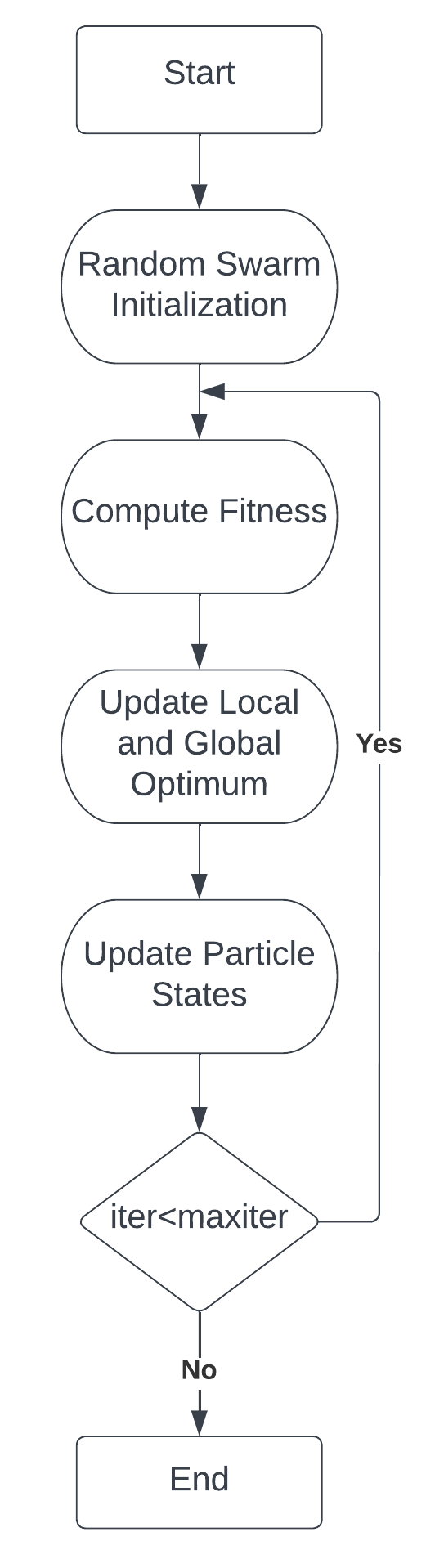}
\caption{Schematics related to PSO flowchart \cite{bib18}.}\label{fig9}
\end{figure*}

\subsection{Genetic Algorithm}\label{subsec9}

\begin{figure*}[ht]
\centering
\includegraphics[width=0.17\textwidth]{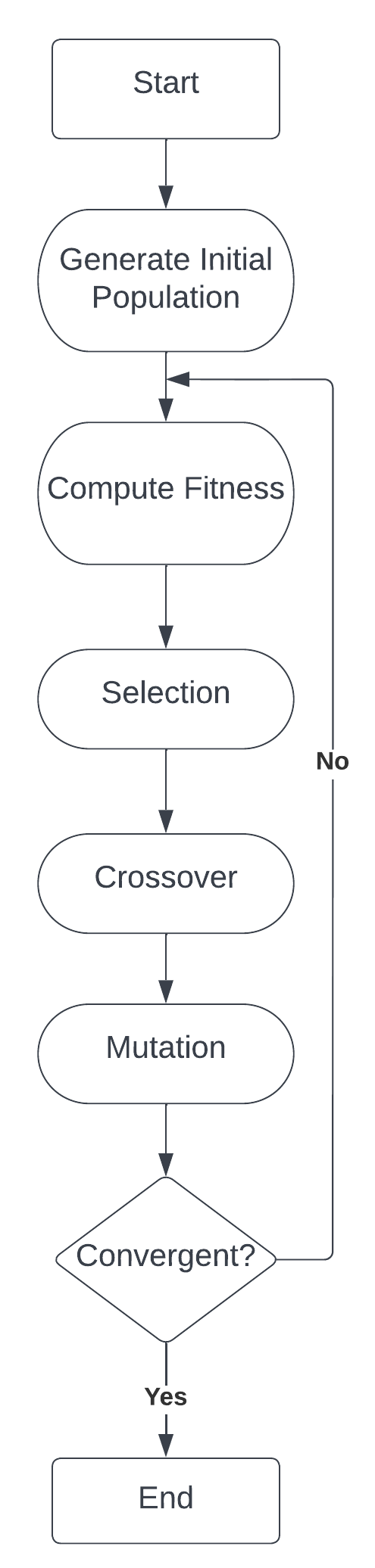}
\caption{Schematics and corresponding GA flowchart \cite{bib20}.}\label{fig10}
\end{figure*}

A genetic algorithm (GA) as used in the papers \cite{bib19}, \cite{bib43}, \cite{bib42}, \cite{bib41}, \cite{bib40} is a search heuristic based on Charles Darwin's idea of natural selection. This algorithm is modeled after the said process, in which the fittest individuals are chosen for reproduction to generate offspring for the following generation. The natural selection process begins with the selection of the fittest individuals from a population. The generated offspring inherit the qualities of their parents and are passed on to the next generation. If parents are more fit, the generated offspring will have a higher fitness value than their parents and have a greater chance of survival. This technique is repeated indefinitely until a generation of the fittest individuals is discovered. The selection phase's goal is to choose the fittest individuals and allow them to pass on their genes to the next generation. Individuals with high fitness have a better probability of being chosen for reproduction. The most important phase in a genetic algorithm is crossover. For each pair of parents to be mated, a crossover point is picked at random from within the genes. Some of the genes in a particular newly generated offspring can be susceptible to a mutation with a low random chance. Mutation happens to preserve population variety and to prevent premature convergence. If the population has converged, the algorithm will terminate. The genetic algorithm has then delivered a collection of solutions to our objective function, the underlying functionalities are mentioned in the figure \ref{fig10}.

\subsection{Differential Evolution}\label{subsec10}
Differential evolution (DE) \cite{bib21} is a method in evolutionary computing that optimizes a problem by iteratively attempting to enhance a candidate solution concerning a given measure of quality. These approaches are usually referred to as metaheuristics since they make little or no assumptions about the issue to be solved and may search very huge areas of potential solutions. Metaheuristics such as DE, on the other hand, do not guarantee an optimum solution. DE is used to optimize multidimensional real-valued functions but does not employ the gradient of the issue being optimized, which means DE does not require the optimization problem to be differentiable, as gradient descent and quasi-newton techniques do. As a result, DE may be applied to optimization problems that are not even continuous, vary over time, and have constituent noise \cite{bib22}.

DE optimizes a problem by storing a population of candidate solutions and producing new ones by merging old ones using simple equations, and then keeping whatever candidate solution has the greatest score or fitness on the optimization issue at hand. The optimization problem is regarded as a black box that only provides a measure of quality given a candidate solution, and the gradient is therefore unnecessary.

\subsection{Simulated Annealing}\label{subsec11}

Simulated annealing (SA) as depicted in the papers \cite{bib23}, \cite{bib45} can be explained as a probabilistic procedure for approximating the global optimum of a given function. Specifically, it is a metaheuristic to approximate global optimization in a considerable search space for an optimization problem \cite{bib44}. The algorithm's name is derived from metallurgy's annealing procedure, which involves heating and controlled cooling of a material to change its physical characteristics. Both are material properties that are determined by their thermodynamic free energy \cite{bib24}. It is often used when the search space is discrete, commonly seen in the boolean satisfiability problem, the traveling salesman problem, job-shop scheduling, and protein structure prediction. Simulated Annealing may be preferable to exact algorithms such as gradient descent or branch.
 
 \subsection{Basin-Hopping}\label{subsec12}

Basin-hopping \cite{bib25} is a global optimization approach in applied mathematics that iterates by randomly perturbing coordinates, conducting local optimization, and accepting or rejecting new coordinates depending on a minimized function value. It is a particularly helpful approach for global optimization in very high-dimensional landscapes, such as determining the minimal energy structure for molecules. The algorithm is greatly influenced by Monte-Carlo Minimization \cite{bib26} and can be described as an incremented and improved version of the former.

\subsection{Dual Annealing}\label{subsec13}

Dual Annealing (DA) available from \cite{bib27} is a stochastic global optimization process that is an implementation of the traditional simulated annealing (CSA) algorithm \cite{bib24}. It is based on the generalized simulated annealing (GSA) \cite{bib28} technique which was previously mentioned. It combines the annealing schedule (the pace at which the temperature lowers during algorithm iterations) of "fast simulated annealing" (FSA) with the probabilistic acceptance of an alternative statistical process known as "Tsallis statistics." As a result, it is intended for objective functions with a nonlinear response surface. It falls under the paradigm of stochastic optimization methods, which means that it employs randomization throughout the search process and that each iteration of the search may yield a different solution.

\subsection{Simplicial Homology Global Optimisation}\label{subsec14}

This subsection provides a summarised description of Simplicial Homology Global Optimisation (SHGO) available at \cite{bib27}, the algorithm utilizes concepts from combinatorial integral homology theory to find sub-domains which are, approximately, locally convex and provides characterizations of the objective function as the algorithm progresses. The SHGO algorithm is appropriate for solving general-purpose NLP and black-box optimization problems to global optimality (low dimensional problems) \cite{bib29}. This class of optimization is also known as CDFO (constrained derivative-free optimization). While most of the theoretical advantages of SHGO are only proven when the function to be optimized is a Lipschitz smooth function. The algorithm is also proven to converge to the global optimum for the more general case where the objective function is non-continuous, non-convex, and non-smooth for the original sampling method \cite{bib29}.


\section{Experiments}\label{sec4}

This section presents an empirical analysis of the mentioned algorithms, the experiments are performed in identical conditions with the same value bounds and cost function, the algorithms were deployed using the SciPy \cite{bib27} and Scikit-Opt used in \cite{bib30} toolkits for a thorough analysis. To evaluate the different optimization strategies, two parameters, namely the fitness value and the best probable execution time are taken.

The geometry of the simplified model is to be optimized during the rocker-bogie optimization is based on the objective function stated in equation (11). 

\begin{equation}
\begin{split}
fitness = w _{1} s \mu^{\ast}  +  w  _{2} 
   \varepsilon _{ \mu } ^{\ast} +   w  _{3}  \big(s-1\big) 
     P \cr + w _{4}c _{lat}  + w _{5} c _{long}    +  w _{6}  
       \big( \varepsilon  _{1}   \big)  + w _{7}c _{traff}   
         \cr + w _{8}z _{rw}  +  w _{9} \theta  _{rover}) 
\end{split}
\end{equation}

\begin{table*}[ht]
\begin{center}
\begin{minipage}{174pt}
\caption{ Performance Metrics Weights for the Optimization Series \cite{bib1}. }\label{tab2}
\begin{tabular}{@{}llllllllll@{}}
\toprule
Weight No. & $w1$  & $w2$ & $w3$ &$w4$ & $w5$ & $w6$ &$w7$ &$w8$ &$w9$ \\
\midrule
Weight   & -2  &  -2  & 2 & 1 & 5 & -3 & 2 & -1 & -1\\
\botrule
\end{tabular}
\end{minipage}
\end{center}
\end{table*}

as stated in Table \ref{tab2}, where $w1$ to $w9$ are the weights. The virtual friction coefficient is denoted by the parameter $\mu^{ \ast }$. The difference in virtual friction coefficient at each location is minimized by reducing the error. 

\begin{table*}[ht]
\begin{center}
\begin{minipage}{\textwidth}
\caption{The Optimization Variables Upper (UB) and Lower (LB) Bound \cite{bib1}.}\label{tab3}
\begin{tabular*}{\textwidth}{@{\extracolsep{\fill}}llllllllllll@{\extracolsep{\fill}}}
\toprule
Variables & $X_r$  & $Y_r$ & $Z_r$ &$\gamma_{rb}$ & $X_{b1}$ & $Y_{b1}$ & $X_{b2}$ &$Y_{b2}$ &$L_{rb}$ & $j$ & $c$  \\
\midrule
LB(mm)   & 100  &  100  & 100 & 90[deg] & 100 & 100 & 100 & 100  & 20 & 50 & 1[-]\\
UB(mm)   & 500  &  300  & 200 & 180[deg] & 200 & 300 & 200 & 300 &  100 & 500 & 5[-]\\
\botrule
\end{tabular*}
\end{minipage}
\end{center}
\end{table*}

The fitness equation utilized is similar to that in the original study, with the exception that one of the Load equalization factors, $\epsilon 1$, is removed owing to an assumption that $X_{b1}$ and $X_{b2}$ are equal to $X_{b}$, as described in the relevant literature \cite{bib1}.

\begin{figure*}[ht]
\centering
\includegraphics[width=0.46\textwidth]{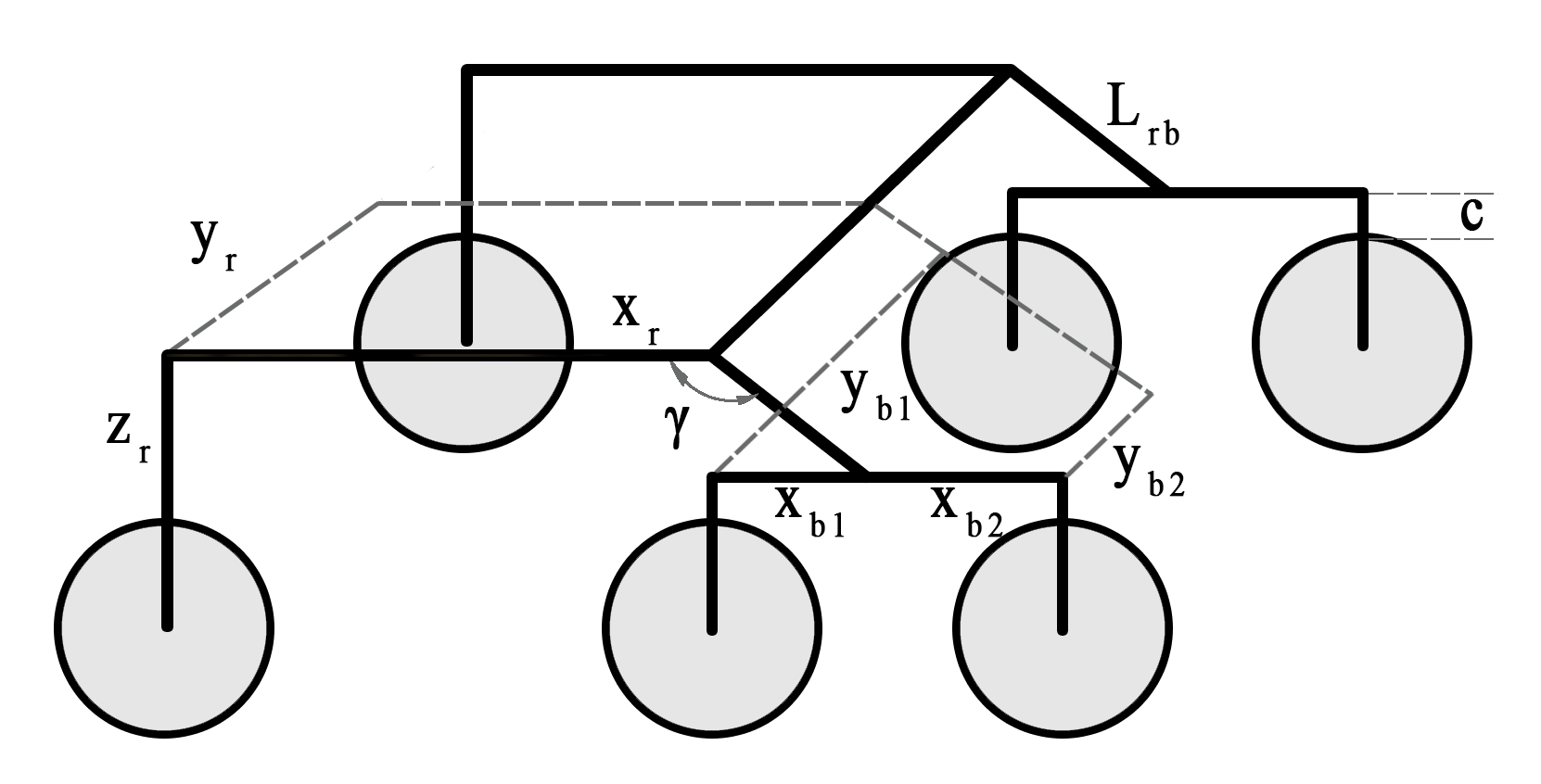}
\caption{All Geometrical Parameters are shown in the Rocker-Bogie Geometry Model.}\label{fig11}
\end{figure*}

The weight of each wheel is reduced, resulting in identical working conditions for everyone. P stands for power consumption and is proportional to it. Traction should be maximised on rocky terrain, while power consumption should be kept to a minimum on smooth ground. This is ensured by the switching function $s$, which is expressed as 

\begin{equation}
\displaystyle   {s =\begin{cases}1 & , \mathrm{if\, max}  \big( \mid  \alpha  \mid \big)  > C \\0 & ,\mathrm{otherwise} 
\end{cases}}
\end{equation}
Where $\alpha_i$ is the degree of inclination of the terrain at each wheel-terrain contact point.$C$ is an arbitrary threshold that determines whether the terrain is deemed rough or benign. In addition, $c_{lat}$, $c_{long}$, and $c_{traff}$  are parameters that are high (= 1000) if lateral stability, longitudinal stability, and trafficability are provided, and zero otherwise. In order to guarantee load equalisation, parameters $\epsilon _1$ and $\epsilon _2$ should be decreased. Sinkage $z_{rw}$ and $\theta_{rover} $ pitch rover are kept to a minimum. As indicated in Fig\ref{fig1} and Fig\ref{fig11} , the geometrical parameters of the rocker-bogie model, as well as the gear ratio $j$  of the differential mechanism linking both rocker-bogies, must be optimised.

\section{Results}\label{sec5}

The rigid wheels are 170 mm in diameter $(d_w)$ and 75 mm wide $(b_w)$. The rover is travelling over sand, and the maximum barrier height to be climbed is$h$ = 170 mm, which is equal to the diameter of the wheel. As indicated in Table \ref{tab4}, all performance metrics are weighted.

\begin{sidewaystable}
\sidewaystablefn
\begin{center}
\begin{minipage}{\textheight}
\caption{The algorithms follow their common abbreviations, the execution time depicts the total computational time taken for one iteration of the algorithm. The fitness values are depicted based on their variable input seed, indicating a diverse empirical analysis.}\label{tab4}
\begin{tabular*}{\textheight}{@{\extracolsep{\fill}}lccccc@{\extracolsep{\fill}}}
\toprule
Algorithm & Execution Time   & Fitness Value
 & Fitness Value & Fitness Value
 & Fitness Value
\tabularnewline
& (Seconds)  & 
(Mean) & (Random) & 
(Upper Bound) & 
(Lower Bound)  \\
\midrule
PSO   & 0.491  &  -  & 714 & - & - \\
SA   & 2.395  & 759  & 759 & 760 & 757 \\
GA   & 4.756  &  -   & 670 & - & - \\
DE   & 28.995  &  732 & 722 & 715 & 732 \\
BH   & 8.624  & -109  & 102 & 634 & -150 \\
DA   & 0.317  &  657  & 606 & 684 & 669 \\
SHGO   & 247.661  &  -  & 407 & - & - \\
\botrule
\end{tabular*}
\end{minipage}
\end{center}
\end{sidewaystable}

It can be concluded that using PSO did give superior performance when compared with the genetic algorithm, showcasing an increase in the fitness value and extreme computational inexpensiveness with an 89.67\% decrease in execution time. The most computationally expensive algorithm turned out to be SHGO while DA illustrated the opposite. For analyzing the performance of an optimization algorithm, based on the permissibility of an input seed, four strategies are employed, namely:

\begin{enumerate}[1.]
\item Initializing using the Lower Bound.

\item Initializing using the Upper Bound.

\item Initializing using the Boundary Mean.

\item Initializing using a random data point between the boundary conditions. 

\end{enumerate}

The graphical description of the said inquiry is mentioned below along with the related algorithm abbreviations and the obtained fitness value.

Through Table \ref{tab4}, a multitude of information can be obtained, the algorithm SA was robust to the input seed value and gave an identical performance on each iteration, similar inferences can also be obtained for DE and DA. Basin Hopping on the other hand depicted utmost variance, where two input scenarios failed to find the minima and gave a negative fitness value, hence questioning the robustness and the algorithms susceptibility. 
For further assessing and validating the performance of each algorithm concerning evolutionary iterations, multiple graphs were obtained. Each graph shows a comparative analysis based on the value obtained by the best individual in a generation or the final value at a particular iteration. The term generation is mainly applicable to PSO and GA, however for algorithms that do not deploy multiple agents, the term iteration is also depicted as a generation. Each methodology is tested for 100 generations and charts are obtained for each feasible metric and parameter. The performance metrics associated with alpha or the angle of inclination, are avoided due to the constituent randomness and its effect on the graphical description. The final values for the relevant performance metrics and algorithms are mentioned below in Table \ref{tab5}.

\begin{table}[ht]
\begin{minipage}{174pt}
\caption{ Final converged values of multiple heuristic strategies for performance metrics.}\label{tab5}%
\begin{tabular}{@{}llll@{}}
\toprule
Performance Metric & PSO & GA & SA \\
\midrule
Power & $ 74.8x10^9$ & $5.14x10^9$ & $6.38x10^9$ \\
$\epsilon 1$ & $-211.86$ & $-162.55$ & $-211.02$ \\
Theta Rover & $175.22$ & $167.98$ & $171.48$ \\
\botrule
\end{tabular}
\end{minipage}
\end{table}

\begin{figure}[ht]
\centering
\includegraphics[width=0.50\textwidth]{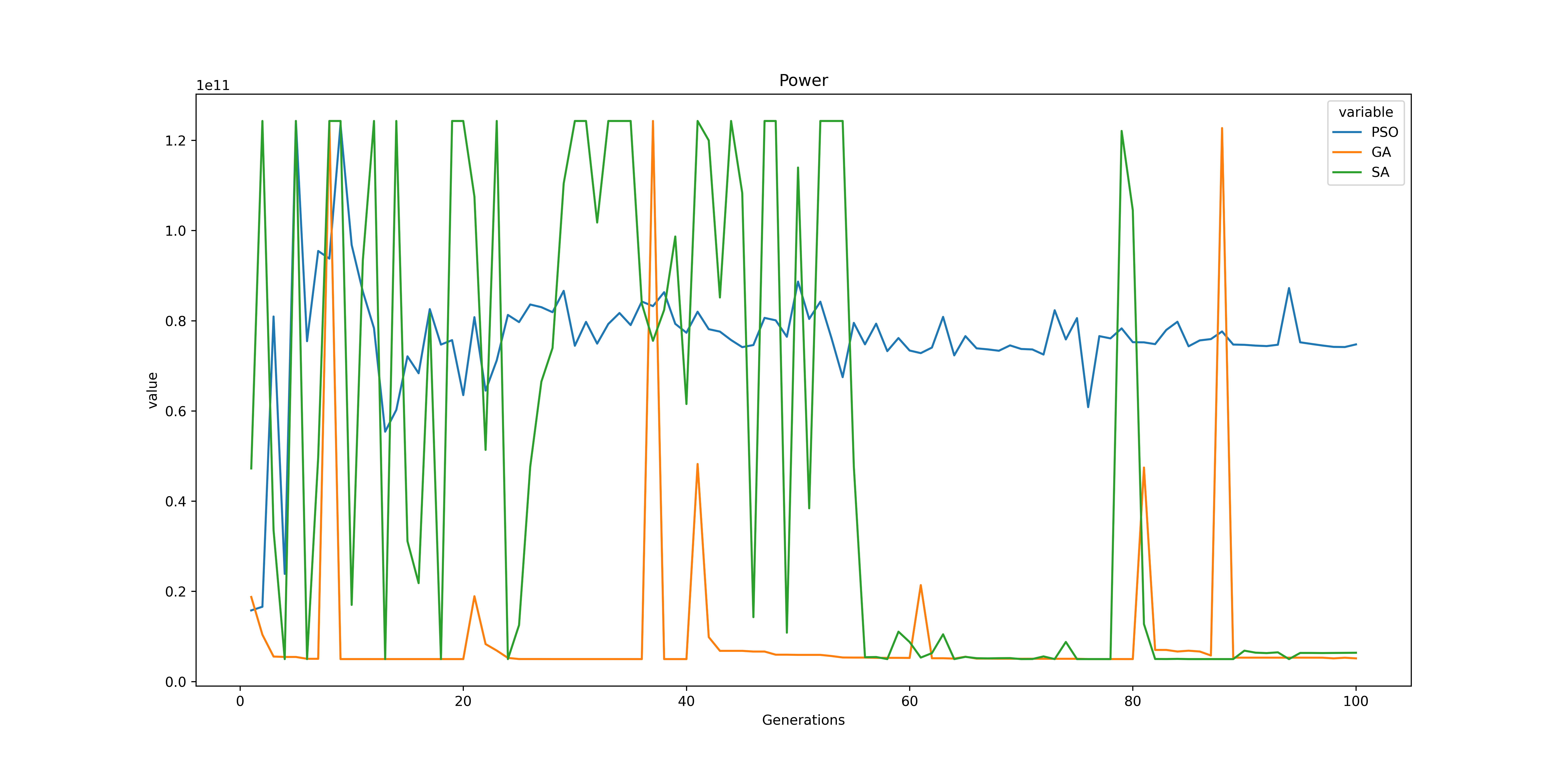}
\caption{Multi-line plot for the Power performance metric.}\label{fig12}
\end{figure}
As the paper’s initial motivation implicated the utility of PSO as a replacement for GA, the superior performance of SA for the objective function, and for maintaining a concise display of information these three algorithms are used. From Fig\ref{fig12} it can be inferred that for the metric Power the final generational entity was similar for GA and SA with a 10.07\% difference, however, PSO showed an 87.14\%, 84.28\% disparity respectively.

\begin{figure}[ht]
\centering
\includegraphics[width=0.50\textwidth]{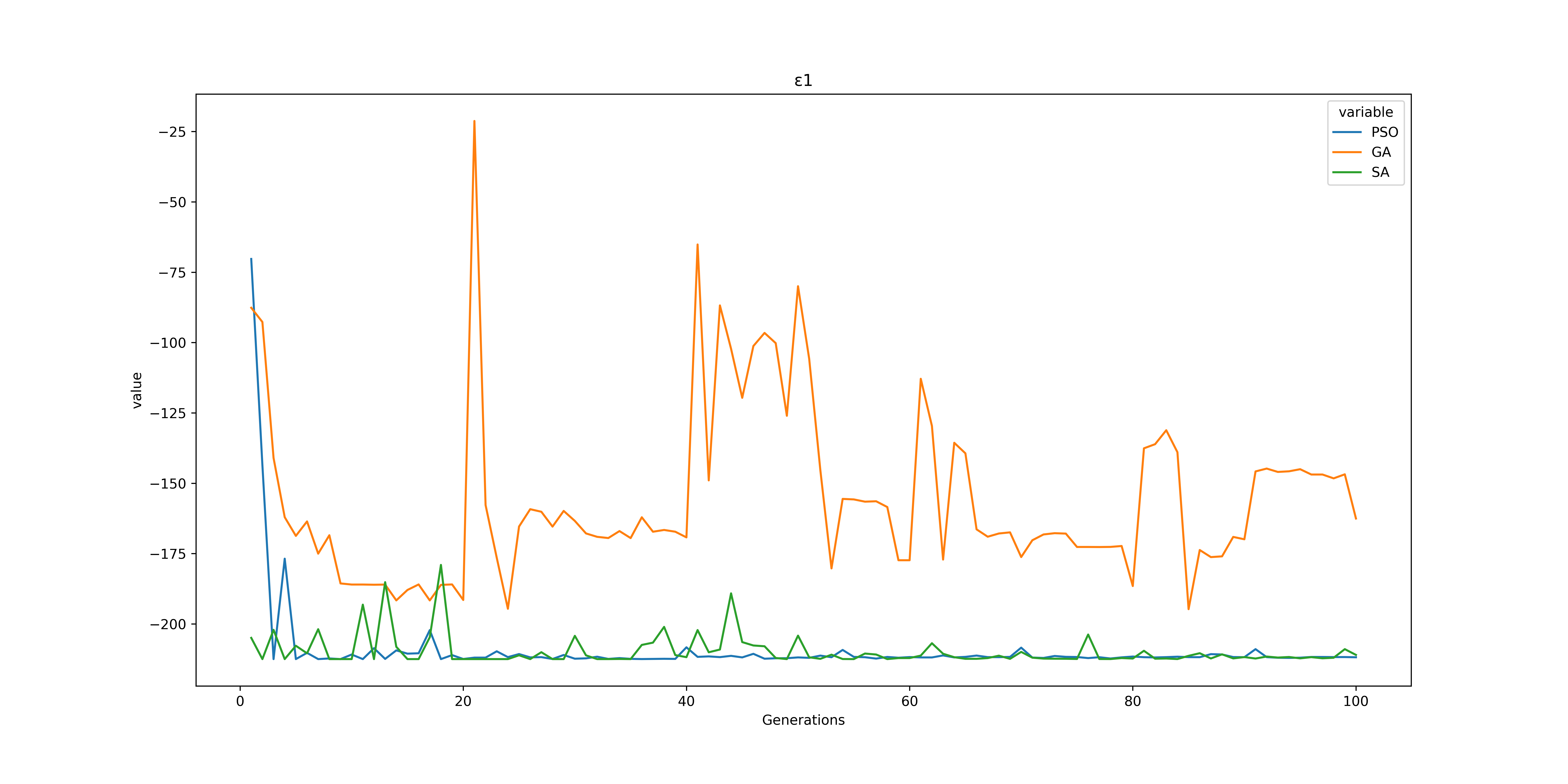}
\caption{Multi-line plot for the $\epsilon 1 $ performance metric.}\label{fig13}
\end{figure}

The above-mentioned graph in Fig\ref{fig13} implicates a similar performance of SA, PSO with only a 0.198\% difference, however, when these algorithms were compared to the genetic equivalent, a 12.97\%, 13.17\% disparity was obtained. Considering the -3 weight value associated with Epsilon1, the impact of PSO, SA, and GA on the fitness was ‘635.58’, ‘633.06’, and ‘487.65’ respectively. 

\begin{figure}[ht]
\centering
\includegraphics[width=0.50\textwidth]{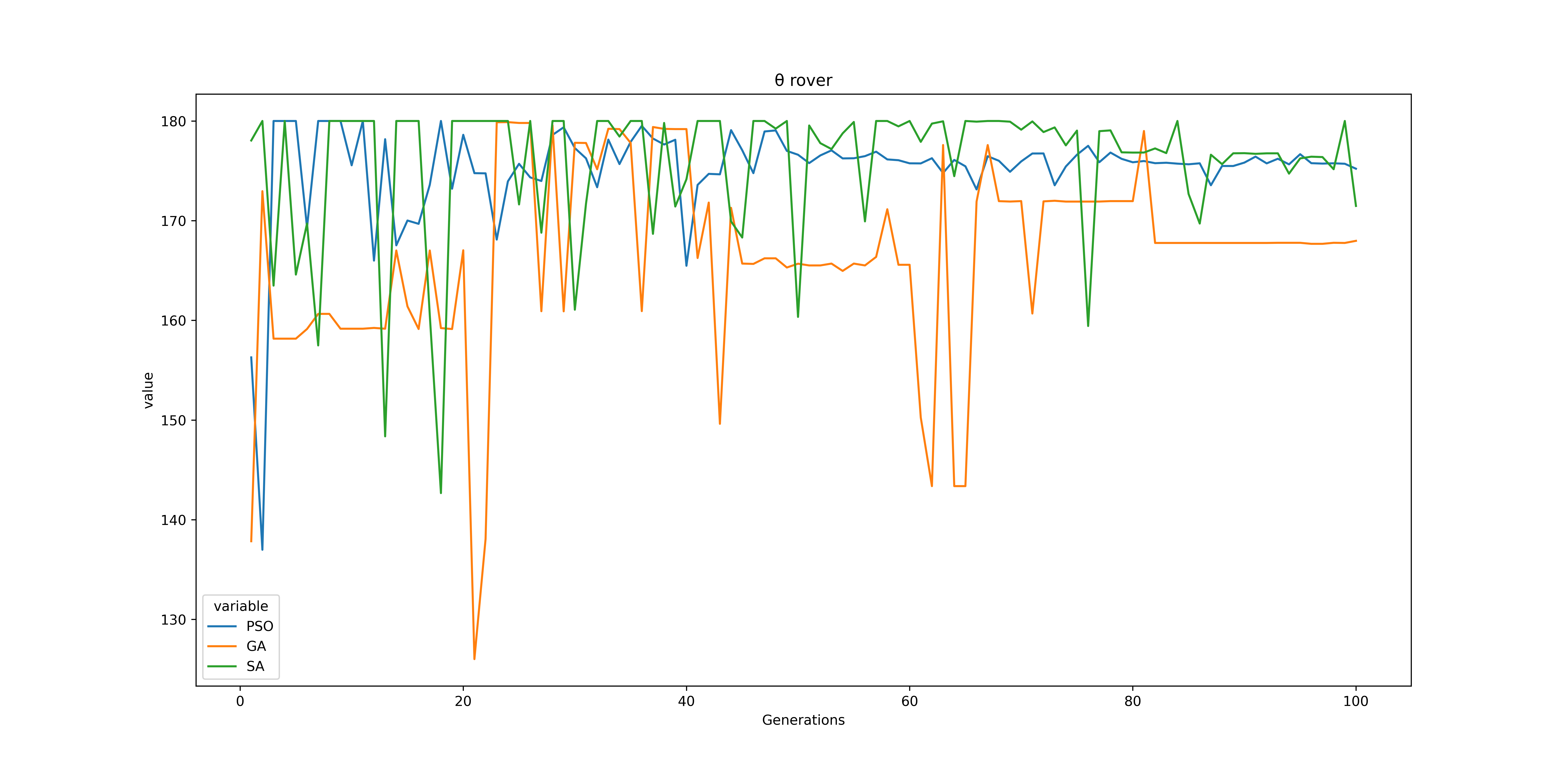}
\caption{Multi-line plot for the $\theta_{rover}$ performance metric.}\label{fig14}
\end{figure}

The empirical performance analysis for this metric was extremely consistent across the three constituent algorithms, depicting negligible variance in comparison with the other relevant metrics. The above can also be inferred by Fig\ref{fig14}.
The individual components or integral parameters were also thoroughly assessed by using a graphical description. The optimal converged values by the three highly relevant algorithms, PSO, GA, and SA are mentioned below in Table \ref{tab6}.

\begin{sidewaystable}
\sidewaystablefn
\begin{center}
\begin{minipage}{\textheight}
\caption{The algorithms follow their common abbreviations, the execution time depicts the total computational time taken for one iteration of the algorithm. The fitness values are depicted based on their variable input seed, indicating a diverse empirical analysis.}\label{tab6}
\begin{tabular*}{\textheight}{@{\extracolsep{\fill}}lccc@{\extracolsep{\fill}}}
\toprule
Geometrical Parameters & PSO & GA & SA  \\
\midrule
$X_{r}$ & 100.42 & 183.31 & 100.01 \\
$Y_{r}$ & 297.42 & 159.24 & 299.98 \\
$Z_{r}$ & 193.80 & 150.78 & 100.00 \\
$\gamma_{rb}$ & 175.21 & 167.98 & 171.48 \\
$X_{b}$ & 100.01 & 102.21 & 100 \\
$Y_{b1}$ & 199.56 & 157.91 & 199.80 \\
$Y_{b2}$ & 290.34 & 165.99 & 297.20 \\
c & 61.59 & 76.57 & 100 \\
$L_{rb}$ & 500.00 & 492.94 & 499.86 \\
j & 3.87 & 1.02 & 1.13 \\
\botrule
\end{tabular*}
\end{minipage}
\end{center}
\end{sidewaystable}

\begin{figure}
\centering
\includegraphics[width=0.50\textwidth]{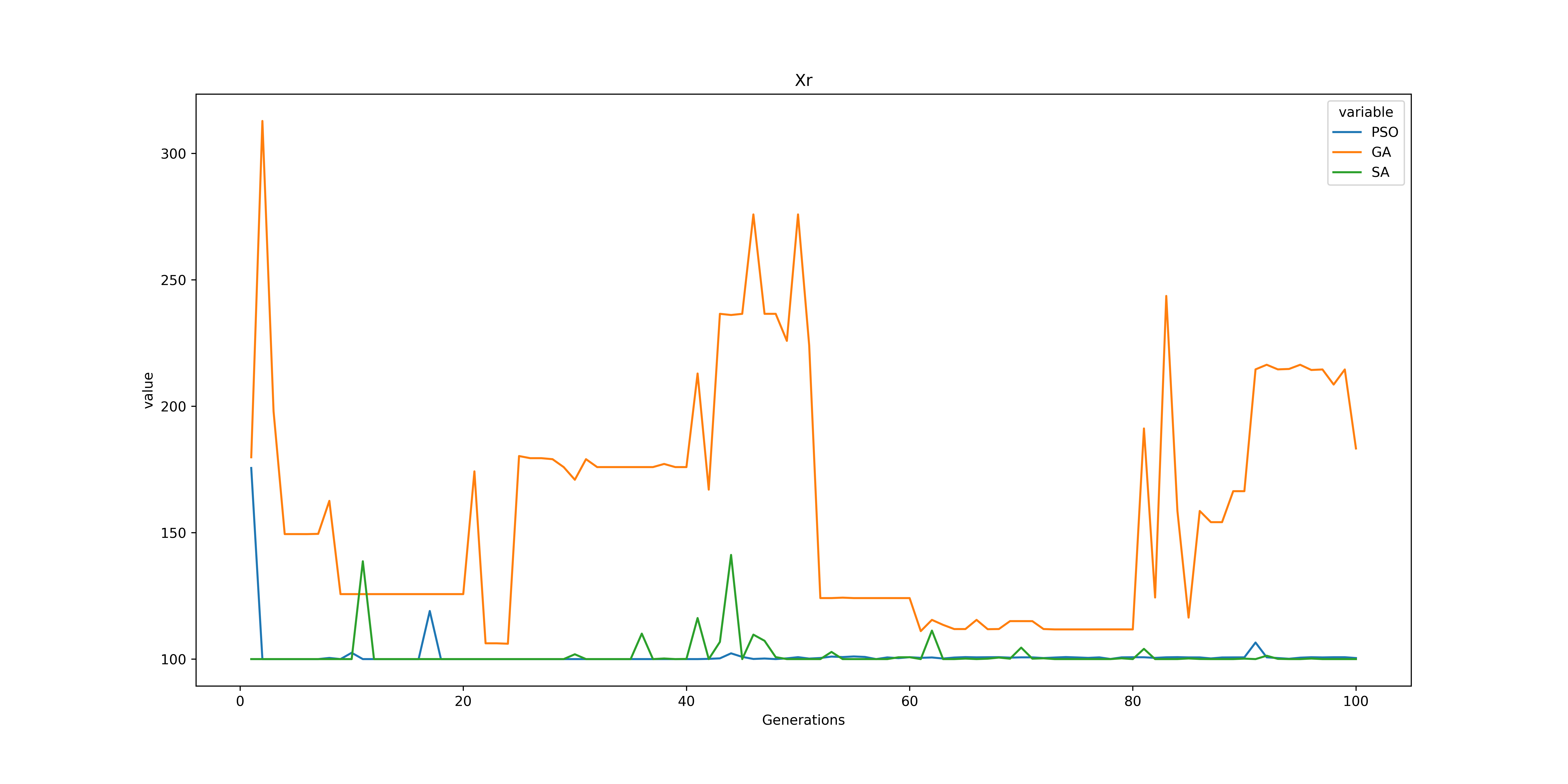}
\caption{ $X_r$ through 100 scaled generation}\label{fig15}
\end{figure}

\begin{figure}
\centering
\includegraphics[width=0.50\textwidth]{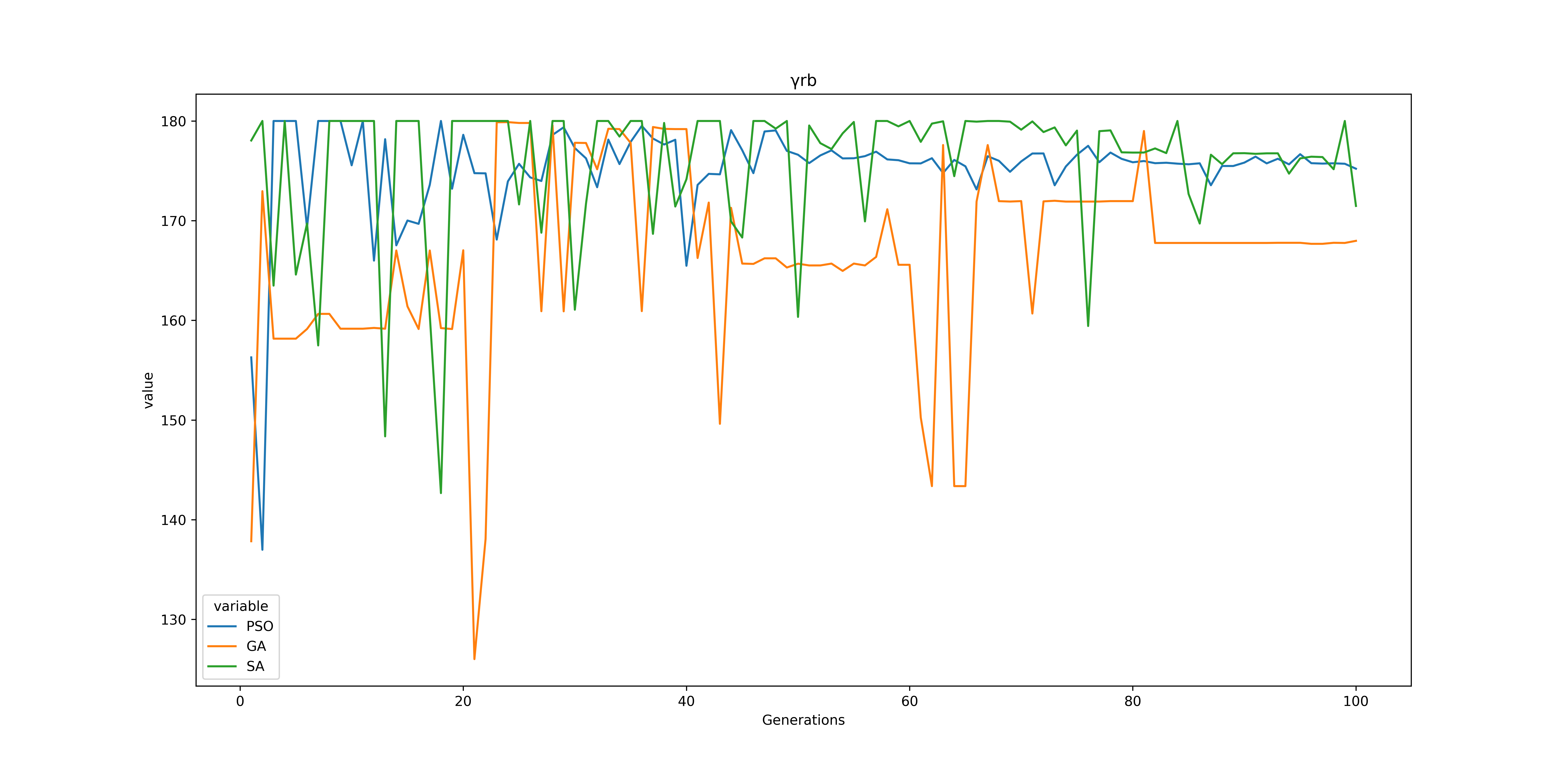}
\caption{$\gamma_{rb}$ through 100 scaled generations, a clear similarity between the final optimal values can be observed}\label{fig16}
\end{figure}

\begin{figure}
\centering
\includegraphics[width=0.50\textwidth]{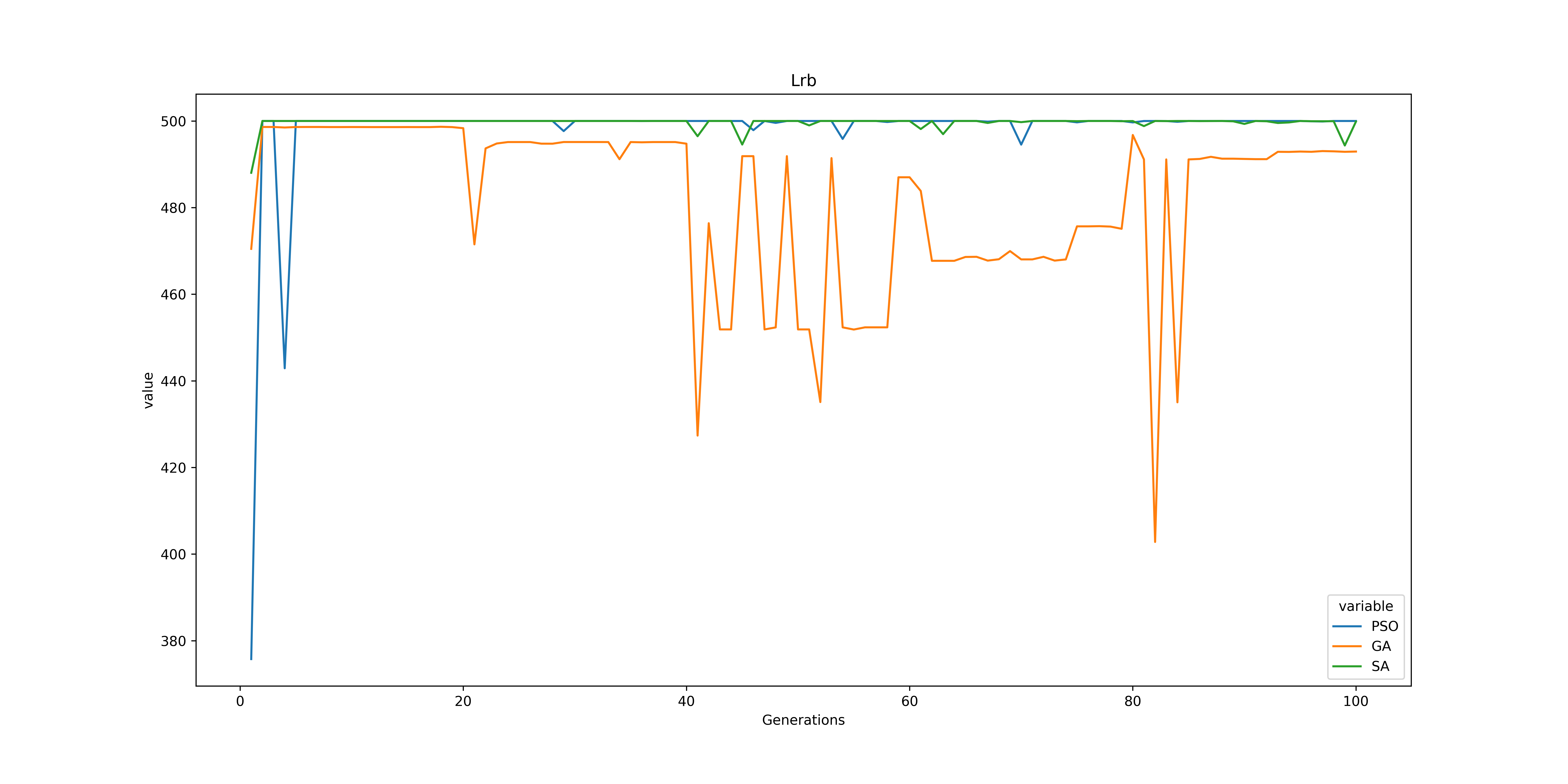}
\caption{Corresponding graph for $L_{rb}$, the following test offers deeper insight and performance-based correlation between the three algorithms.}\label{fig17}
\end{figure}

\begin{figure}
\centering
\includegraphics[width=0.50\textwidth]{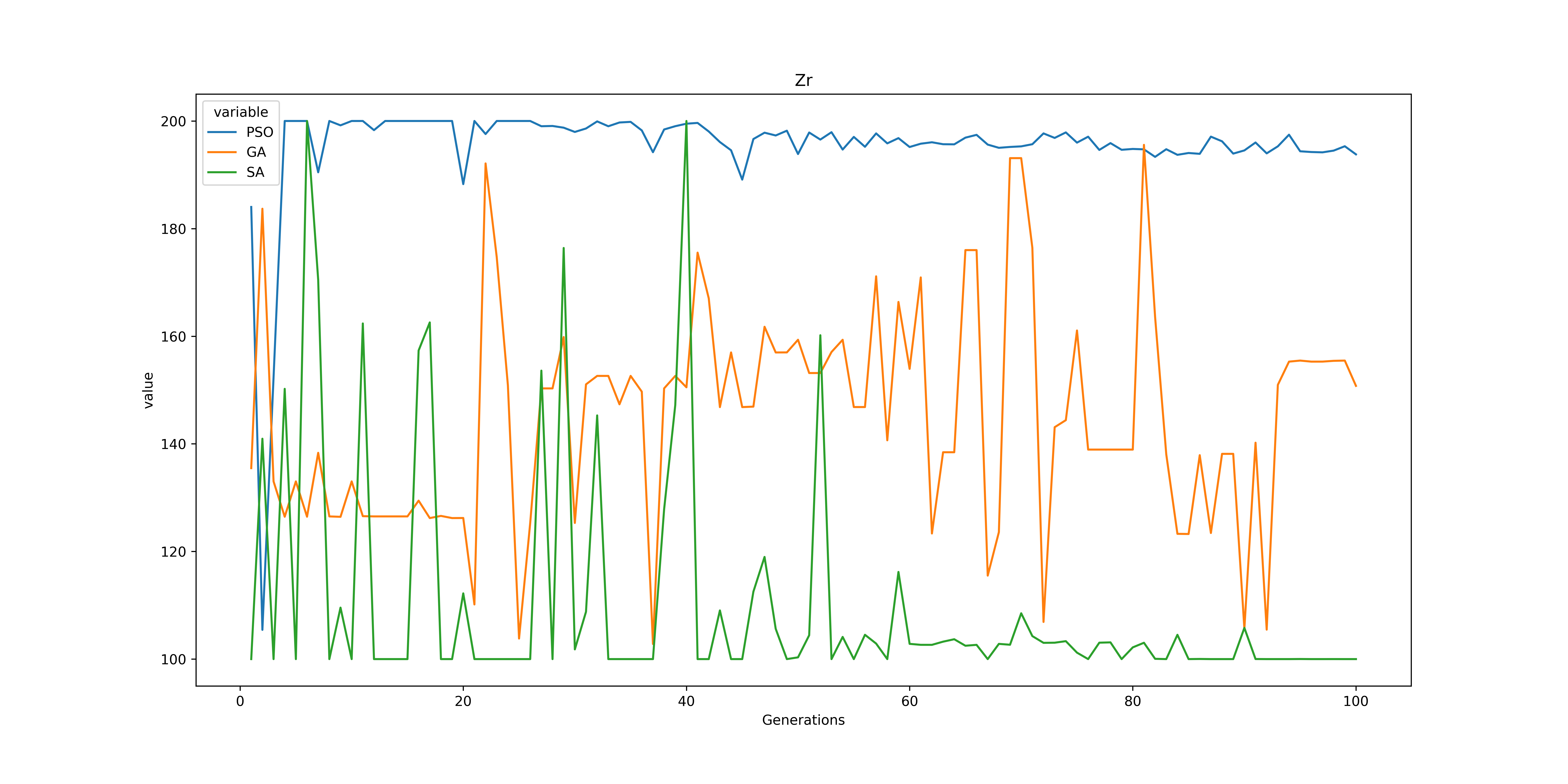}
\caption{ Graphical description of $Z_{r}$, the only geometrical parameter which depicted a different performance for all the heuristic approaches.
}\label{fig18}
\end{figure}

When the performances of various parameters were plotted against a total of 100 scaled generations, a multitude of parameters depicted similar value trends for all the heuristics, however, the parameters $X_b$ , $\gamma_{rb}$ , $Z_r$, and $L_{rb}$ stood out the most. The aforementioned statement can be validated by assessing the figures \ref{fig15}, \ref{fig16}, \ref{fig17}, \ref{fig18}.

From the corresponding graph illustrated in Fig\ref{fig15} for the $X_r$ parameter, multiple similarities can be obtained for SA and it's swarm equivalent as they converged to the final value in extremely few, initial generations, however, GA failed to  offer performance-based optimality. The values ranged from 100 to 350 which is analogous to the aforementioned constraints, by being a subset of [100,500]. 

Only the values associated with $L_{rb}$, and $\gamma_{rb}$ were obtained to be identical or similar across all the heuristic approaches, with negligible disparities and differences when validated against other parameters.

\section{Conclusion}\label{sec6}

This paper outlines the design of a quasi-static rocker-bogie suspension system for an extraterrestrial rover, as well as the identification of numerous performance indicators that must be improved if the rover is to fulfill its jobs and achieve its objectives as efficiently as feasible. Through thorough empirical analysis, we were able to conclude the use of Simulated Annealing concerning other state-of-the-art approaches. Based on our initial assumption and motivation regarding the use of PSO, we thoroughly assess the said algorithm and obtain superior results in comparison to the baseline Genetic Algorithm. Tests were conducted to check the modality and nature of heuristic approaches on varied inputs from the possible hyperspace space defined by the constraints, these tests corresponded to the robustness estimation and also offer insights on its corresponding utility. Various performance metrics, geometrical parameters, and their related historical values through 100 scaled generations or iterations are plotted to accurately understand the trends associated with the implementations. The best performing heuristic approach was obtained to be SA with a fitness of 760, which was relatively superlative when compared to other algorithms, and offered a consistent performance across variable input seeds and individual performance metrics. For the future we aim to leverage various optimization and minimization algorithms as an added layer to standard swarm intelligence algorithms, further testing the associated input seed functionalities.\\

\textbf{Conflict of Interest} : The authors have no conflict of interest to declare.\\        

\textbf{Funding}: There is no funding source.\\

\textbf{Ethical approval}: This article does not contain any studies with human participants or animals performed by any of the authors.\\

\textbf{Informed consent}: Informed consent was obtained from all individual participants included in the study.\\

\end{document}